\pdfoutput=1
\documentclass[a4paper, 9pt, twocolumn]{article} % For LaTeX2e

  \usepackage{graphicx}
  \usepackage{mathrsfs}
  \usepackage{subcaption}
  \usepackage{array}
  \usepackage{amsfonts}
  \usepackage{amsmath}
  \usepackage{hyperref}
  \usepackage{eurosym}
  \usepackage{multirow}
  
  \usepackage[top=2cm, bottom=2cm, left=1.7cm, right=1.7cm]{geometry}
  
  \usepackage[T1]{fontenc}

  \usepackage{fancyhdr}
  \usepackage{etoolbox}
  \AtBeginEnvironment{tabular}{\scriptsize}

  \usepackage{wrapfig}

  \usepackage[dvipsnames]{xcolor}
  \usepackage{xcolor}
  \usepackage{amsmath}
  \usepackage{gensymb}
  \usepackage{multirow}
  \usepackage{array}
  \usepackage{booktabs}
  \usepackage{graphicx}
  \usepackage{url}
  \usepackage{makecell}

  \newcommand{\corrA}[1]{#1}

\begin{document}
%   \fontfamily{lmss}\selectfont

  \twocolumn[{\centering{\Huge
  Ionospheric activity prediction using convolutional recurrent neural networks
  \par}\vspace{3ex}
	{\Large Alexandre~Boulch$^1$,
  No\"elie~Cherrier$^2$
  and~Thibaut~Castaings$^3$
  \par}\vspace{2ex}
  $^1$ ONERA, The French Aerospace Lab, Chemin de la Huni\`ere, 91123 Palaiseau, France\\
  $^2$ No\"elie Cherrier works for Irfu, CEA, Universit\'e Paris-Saclay, 91191, Gif-sur-Yvette, France\\
  $^3$ Thibaut Castaings works for BCG (The Boston Consulting Group) in Paris office
  \par\vspace{4ex}
  }
  {\noindent\makebox[\linewidth]{\rule{\linewidth}{1pt}}}
  {\centering\bfseries Abstract\par}
  \smallbreak
  The ionosphere electromagnetic activity is a major factor of the quality of satellite telecommunications, Global Navigation Satellite Systems (GNSS) and other vital space applications. 
  Being able to forecast globally the Total Electron Content (TEC) would enable a better anticipation of potential performance degradations.
  A few studies have proposed models able to predict the TEC locally, but not worldwide for most of them.
  Thanks to a large record of past TEC maps publicly available, we propose a method based on Deep Neural Networks (DNN) to forecast a sequence of global TEC maps consecutive to an input sequence of TEC maps, without introducing any prior knowledge \corrA{other than Earth rotation periodicity}. 
  By combining several state-of-the-art architectures, the proposed approach is competitive with previous works on TEC forecasting while predicting the TEC globally.
  
  \medbreak
  \textbf{Keywords:} Sequence prediction, neural network, forecast, ionosphere, TEC, deep learning, CNN, RNN
  
  {\noindent\makebox[\linewidth]{\rule{\linewidth}{1pt}}}
  \par\vspace{2ex}
  ]
  
  %%%%%%%%%%%%%%%%%%%%%%%%%%%%%%%%%%%%%%%%%%%%%%%%%%%%%%%%%%%%%%%%%%%%%%%%%%%%%%%%%%%%%%%%%
  %%%%%%%%%%%%%%%%%%%%%%%%%%%%%%%%%%%%%%%%%%%%%%%%%%%%%%%%%%%%%%%%%%%%%%%%%%%%%%%%%%%%%%%%%
  %   _____ _   _ _______ _____   ____  
  %  |_   _| \ | |__   __|  __ \ / __ \ 
  %    | | |  \| |  | |  | |__) | |  | |
  %    | | | . ` |  | |  |  _  /| |  | |
  %   _| |_| |\  |  | |  | | \ \| |__| |
  %  |_____|_| \_|  |_|  |_|  \_\\____/ 
  %%%%%%%%%%%%%%%%%%%%%%%%%%%%%%%%%%%%%%%%%%%%%%%%%%%%%%%%%%%%%%%%%%%%%%%%%%%%%%%%%%%%%%%%%
  %%%%%%%%%%%%%%%%%%%%%%%%%%%%%%%%%%%%%%%%%%%%%%%%%%%%%%%%%%%%%%%%%%%%%%%%%%%%%%%%%%%%%%%%%

  \section{Introduction}\label{sec:introduction}
  
  The ionosphere is the ionized upper part of the atmosphere.
  This atmosphere layer is the first to receive solar wind and radiations, which results in an ionization process that has a significant influence on a wide range of human activities using transionospheric radio waves. 
  The impacted activities include telecommunications, positioning services (Global Navigation Satellite Systems, GNSS) and space observation from the ground.
  During periods of high ionospheric activity, the ions induce a modification of the radio wave path across the atmosphere, resulting in noise increase, significant bitrate reduction or positioning errors \cite{dattabarua2010,lee2011}.
  % Todo correction on positioning systems.
  %To counteract the effect of ionization, several methods such as the Klobuchar compensation model~\cite{klobuchar1987ionospheric} may be used. This model estimates the time delay for satellite-to-Earth communication based on the density of ionized particles in the ionosphere, measured in Total Electron Content units.
  
  The Total Electron Content (TEC) is the main measure for the electromagnetic activity of the ionosphere.
  At a specific location, the TEC is the total number of electrons in the atmosphere along a vertical tube of one square meter cross-section.
  It is expressed in TEC Units:
  \begin{equation}
  1\; \text{TECU} = 10^{16}\ el/m^{2}
  \end{equation}
  usually ranging from a few units to a hundred TECU.
  
  % Forecasting
  Forecasting the activity of the ionosphere at global scale, i.e. TEC maps, increases the ability of space borne service users to evaluate for instance future data loss probabilities or positioning error margins.
  % It is even possible that accurate TEC prediction could lead to real time positioning compensation.
  
  \begin{figure}[!t]
  \centering
  \includegraphics[width=\linewidth]{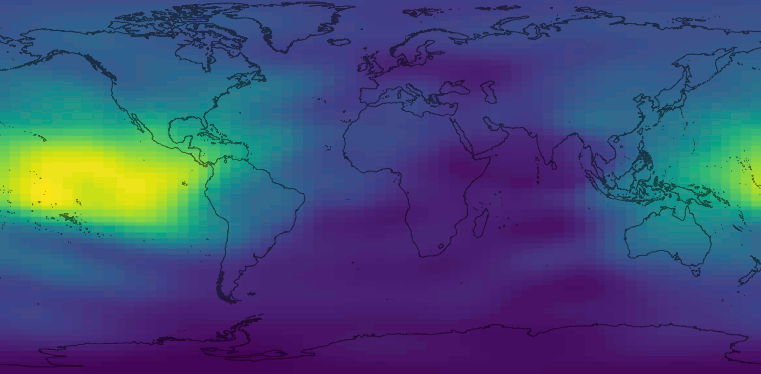}
  \caption{A TEC map. (Dark blue for low values, and yellow for high values.)}
  \label{fig:tec}
  \end{figure}

  \begin{figure*}[!t]
  \centering
  \includegraphics[width=\linewidth]{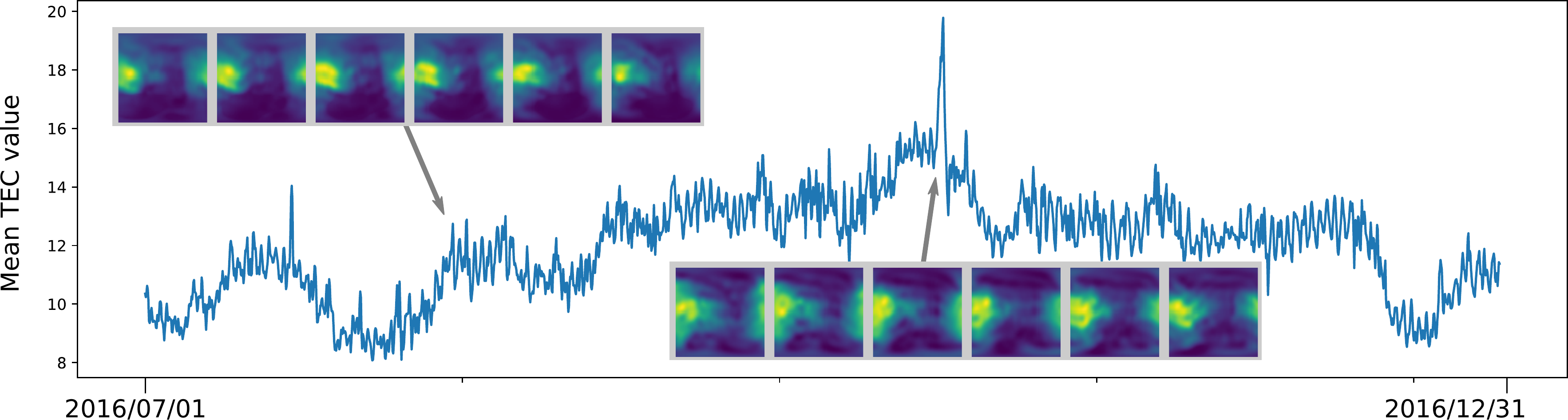}
  \caption{Mean TEC value for all the frames of the test set.}
  \label{fig:test_set}
  \end{figure*}
  
  % Source of the data
  For scientific purpose, the Center for Orbit Determination in Europe (CODE) provides global TEC maps \cite{tulunay2004} with a temporal resolution of two hours.
  An example of TEC map is presented in Figure~\ref{fig:tec}.
  The sunlit face of the Earth corresponds to higher TEC levels (in yellow on the figure) while poles and face at night correspond to low TEC levels.
  Due to the Earth rotation, the solar influence presents a periodicity of 24 hours.
  %Added to the solar factor, the winds in the upper atmosphere play a crucial role in the TEC maps aspect and evolution.
  Moreover, the magnetosphere and the variations of the Earth magnetic field (for example due to solar winds) have an influence on the ionosphere.
  As a result, the TEC maps have a 24-hour periodicity that can be easily mostly removed.
  The inner streams and local perturbations of the ionosphere remain to predict.
  
  % Presentation of the paper
  The objective of this paper is to address the challenge of globally forecasting the TEC map sequence from 2h to 48h ahead of real time.
  This paper is an extension of \cite{Cherrier2017-2} and also widely refers to the conference paper \cite{Cherrier2017}.
  Following these works, we propose a prediction method based on deep convolutional recurrent networks.
  We build different network architectures and compare them for the TEC prediction task.
  Compared to the previous works, we introduce a more robust architecture which is not based on encoder-decoder structure.
  We also do further experiments and analyses to validate the usefulness of such networks for space weather forecasting.
  
  The paper is organized as follows: in Section~\ref{sec:related}, we relate the previous works on TEC prediction and more generally on image sequence forecasting.
  \corrA{
  Section~\ref{sec:data} details the TEC data used in this work and its preprocessing.
  }
  Section~\ref{sec:forecast} presents an overview of the proposed method while recalling the principle of convolutional and recurrent neural networks.
  The details of the network architectures are given in Section~\ref{sec:networks}.
  Experiments are presented in Section~\ref{sec:expe} and perspectives are studied in Section~\ref{sec:perspectives}.

  %%%%%%%%%%%%%%%%%%%%%%%%%%%%%%%%%%%%%%%%%%%%%%%%%%%%%%%%%%%%%%%%%%%%%%%%%%%%%%%%%%%%%%%%%
  %%%%%%%%%%%%%%%%%%%%%%%%%%%%%%%%%%%%%%%%%%%%%%%%%%%%%%%%%%%%%%%%%%%%%%%%%%%%%%%%%%%%%%%%%
  %   _____  ______ _            _______ ______ _____  
  %  |  __ \|  ____| |        /\|__   __|  ____|  __ \ 
  %  | |__) | |__  | |       /  \  | |  | |__  | |  | |
  %  |  _  /|  __| | |      / /\ \ | |  |  __| | |  | |
  %  | | \ \| |____| |____ / ____ \| |  | |____| |__| |
  %  |_|  \_\______|______/_/    \_\_|  |______|_____/
  %%%%%%%%%%%%%%%%%%%%%%%%%%%%%%%%%%%%%%%%%%%%%%%%%%%%%%%%%%%%%%%%%%%%%%%%%%%%%%%%%%%%%%%%%
  %%%%%%%%%%%%%%%%%%%%%%%%%%%%%%%%%%%%%%%%%%%%%%%%%%%%%%%%%%%%%%%%%%%%%%%%%%%%%%%%%%%%%%%%%
  \section{Previous work}
  \label{sec:related}
  
  Several services exist to address TEC forecasting.
  They rely on measurements provided by GNSS ground networks \cite{tulunay2004} and aim at producing global TEC maps.
  CTIPe is an experimental tool implementing complex physics models \cite{millward2001} developed by the US Space Weather Prediction Center that produces global forecasts 30 minutes ahead of real-time.
  In Europe, the ESA Ionospheric Weather Expert Service Center combines products from different national services to provide global and regional 1-hour TEC forecasts.
  However, the records of the input data and forecasts are not published.
  
  A global analytical TEC model has been proposed in \cite{similar_global}, using open source TEC data from the CODE. 
  This model is intended to apply to any temporal range, without relying on a record of TEC values.
  
  The literature provides several methods using time series and statistical methods to predict TEC with various forecasting horizons from a few minutes to several days based on the previous state of the ionosphere.
  Most of these methods \cite{similar6,similar2,similar5,similar3,similar4} provide predictions above specific stations.
  Among these, a few works aim at reconstructing the TEC on a small area \cite{similar_region2,similar_region3} with methods such as Bezier surface-fitting or Kriging.
  Some of them use machine learning, particularly neural networks \cite{similar_region2,similar1,similar_region1}, to infer the model parameters. However, they only focus on local stations.
  Obtaining a regional or global prediction would require one model for each location and interpolation for the areas that are not covered.

  %%%%%%%%%%%%%%%%%%%%%%%%%%%%%%%%%%%%%%%%%%%%%%%%%%%%%%%%%%%%%%%%%%%%%%%%%%%%%%%%%%%%%%%%%
  %%%%%%%%%%%%%%%%%%%%%%%%%%%%%%%%%%%%%%%%%%%%%%%%%%%%%%%%%%%%%%%%%%%%%%%%%%%%%%%%%%%%%%%%%
  %   _____       _______       
  %  |  __ \   /\|__   __|/\    
  %  | |  | | /  \  | |  /  \   
  %  | |  | |/ /\ \ | | / /\ \  
  %  | |__| / ____ \| |/ ____ \ 
  %  |_____/_/    \_\_/_/    \_\
  %%%%%%%%%%%%%%%%%%%%%%%%%%%%%%%%%%%%%%%%%%%%%%%%%%%%%%%%%%%%%%%%%%%%%%%%%%%%%%%%%%%%%%%%%
  %%%%%%%%%%%%%%%%%%%%%%%%%%%%%%%%%%%%%%%%%%%%%%%%%%%%%%%%%%%%%%%%%%%%%%%%%%%%%%%%%%%%%%%%%
  
  \section{TEC data}
  \label{sec:data}

  % The Center for Orbit Determination in Europe (CODE) releases open source TEC data. 
  % The TEC maps have a $5\degree \times 2.5\degree$ resolution on longitude and latitude and 2-hour temporal resolution, covering all latitudes and longitudes. 
  % The value of a pixel corresponds to the vertical TEC at this point.
  % The data is loaded as a sequence of 60 maps (one map every two hours): the first 36 maps (i.e. 3 days) are the inputs for the network, the last 24 maps (i.e. 48 hours) being the prediction targets. 
  % Maps from 1/1/2014 to 5/31/2016 are used for training, and maps from 7/1/2016 to 12/31/2016 for testing. 
  
  % This amount of data is sufficient for the network to understand the context, and more information would lead to an overuse of the neurons of the network (several neurons being dedicated to process out-of-date data).
  % Finally, the frame of reference is changed to Heliocentric, so that the effect of Earth rotation and the Sun heating the ionosphere during the day is compensated.
  
  % For computation purposes, the raw TEC maps are resized to $72 \times 80$ maps and normalized to $[0,1]$ using the highest value of the train set.
  
  %\todo{Extend data part, remove overuse, tecmap 71x73 to 72x72, number of images, is it big data ?}
  
  \corrA{
  The Center for Orbit Determination in Europe (CODE) releases open source TEC data.
  The provided TEC maps are images of size $73 \times 71$ corresponding to about $5\degree \times 2.5\degree$ resolution on longitude and latitude and 2-hour temporal resolution, covering all latitudes and longitudes.
  The value of a pixel corresponds to the vertical TEC at this point.
  These maps are computed using the data from about 200 ground stations including GPS/GLONASS sites of the International GNSS Service (IGS).

  % The data is provided as a continuous service since 2014, leading to a growing amount of data. % on a l'impression que c'est un problème d'avoir trop de données, mais non!
  % Remplacement:
  A huge amount of data has been accumulated since this service started to operate in 2014.
  %In this paper, we aim at showing, that using data previous in a massive data driven approach allows to model physical behavior of ionosphere, producing fast and reliable prediction of the future ionosphere states. 
  % ma proposition:
  In this paper, we take advantage of the high volume of available data to model the physical behavior of the ionosphere using a Deep Learning approach.
  The proposed architectures are capable of forecasting the future states of the ionosphere, accurately and within a short computational time.
  
  \subsection*{Dataset}
  For training and evaluation purposes, we construct a dataset of TEC maps.
  We download TEC maps from 1/1/2014 to 12/31/2016.
  From these images, we build 13000 sequences used for training and testing.
  Each sequence is composed of 60 maps (one map every two hours): the first 36 maps (i.e. 3 days) are the inputs for the network, the last 24 maps (i.e. 48 hours) being the prediction targets.
  Maps from 1/1/2014 to 5/31/2016 are used for training, and maps from 7/1/2016 to 12/31/2016 for testing. 
  The mean TEC level of the test set is presented on Figure~\ref{fig:test_set}.
  The TEC level vary a lot on the 6 months period.
  To illustrate these variation, a few frame are highlighted corresponding to first a typical mean TEC value and second a very disturb sequence.
  In the later, besides a high TEC level, we can observe a lot of high frequencies.
  
  In order to make implementation easier, as we use networks with progressive dimension reduction (see section~\ref{sec:forecast}), we resize the maps to $72 \times 72$.
  The input data is then normalized to $[0,1]$ using the highest value of the train set.
  
  Finally, the last transformation applied to the data is to compensate Earth rotation, i.e. change to Heliocentric coordinate system.
  In this coordinate system, the areas corresponding to sun enlightenment (maximal TEC values in figure~\ref{fig:tec}) are stable through the sequence.
  
  \subsection*{Prediction and evaluation}
  The predictions, i.e. the outputs of the networks, are de-normalized using the inverse normalization process.
  We chose not to clamp to $1$ before de-normalization due to the possibility that the highest value in test set may be higher than the maximum of train set.
  The resulting forecasts are expressed in TEC units.
  }

  %\newpage
  
  %%%%%%%%%%%%%%%%%%%%%%%%%%%%%%%%%%%%%%%%%%%%%%%%%%%%%%%%%%%%%%%%%%%%%%%%%%%%%%%%%%%%%%%%%
  %%%%%%%%%%%%%%%%%%%%%%%%%%%%%%%%%%%%%%%%%%%%%%%%%%%%%%%%%%%%%%%%%%%%%%%%%%%%%%%%%%%%%%%%%
  %   ______ ____  _____  ______ _____           _____ _______ _____ _   _  _____ 
  %  |  ____/ __ \|  __ \|  ____/ ____|   /\    / ____|__   __|_   _| \ | |/ ____|
  %  | |__ | |  | | |__) | |__ | |       /  \  | (___    | |    | | |  \| | |  __ 
  %  |  __|| |  | |  _  /|  __|| |      / /\ \  \___ \   | |    | | | . ` | | |_ |
  %  | |   | |__| | | \ \| |___| |____ / ____ \ ____) |  | |   _| |_| |\  | |__| |
  %  |_|    \____/|_|  \_\______\_____/_/    \_\_____/   |_|  |_____|_| \_|\_____| 
  %%%%%%%%%%%%%%%%%%%%%%%%%%%%%%%%%%%%%%%%%%%%%%%%%%%%%%%%%%%%%%%%%%%%%%%%%%%%%%%%%%%%%%%%%
  %%%%%%%%%%%%%%%%%%%%%%%%%%%%%%%%%%%%%%%%%%%%%%%%%%%%%%%%%%%%%%%%%%%%%%%%%%%%%%%%%%%%%%%%%
  \section{RNN for map forecasting}
  \label{sec:forecast}
  
  As stated in the introduction, the objective is to predict global TEC map sequences given the previous states of the ionosphere.
  The underlying idea of this paper is that a large part of future ionospheric activity can be inferred from its previous states.
  Particularly when looking at the temporal evolution of TEC maps, the main phenomena are continuous, which supports the possibility of predicting the next map sequence.
  This consideration shares similarity with $s$-generalized Markov chains, where the probability of a state at time $t+1$ depends only of a sequence $s$ of previous states.
  
\begin{figure}[!t]
  \centering
  \includegraphics[width=\linewidth]{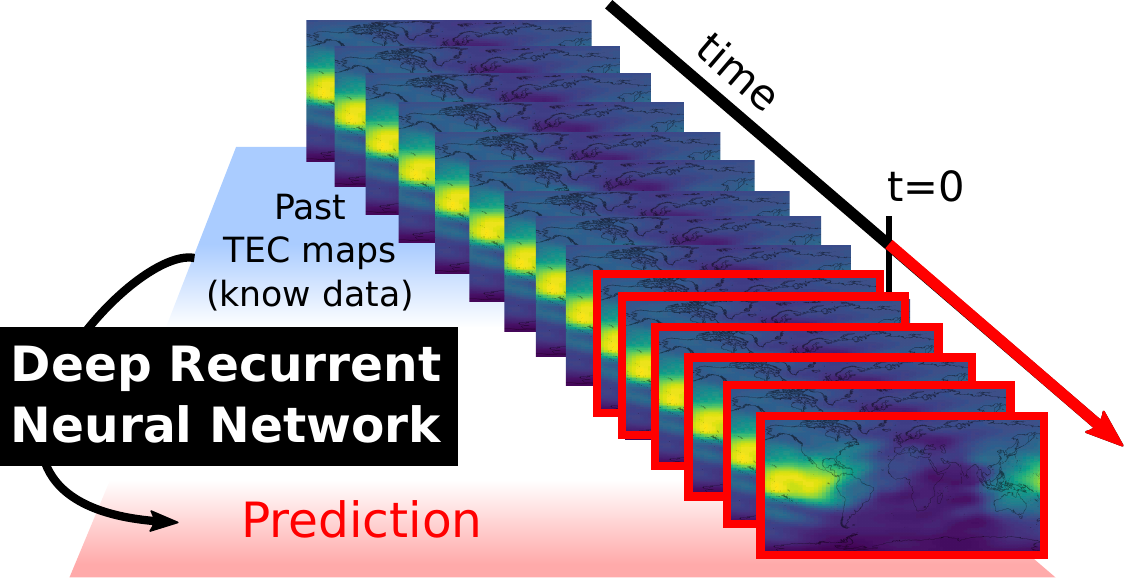}
  \caption{Prediction process.}
  \label{fig:process}
  \end{figure}

  This temporal trend is extracted via Recurrent Neural Networks (RNN), allowing temporal information to flow between processed maps and assist the prediction.
  Figure~\ref{fig:process} presents our prediction scheme.
  We feed a network with a series of input images (the sequence from time $t-s$ to $t$), it outputs a prediction from $t+1$ to $t+p$ ($p$ being the length of the predicted sequence).
  The value of $p$ is up to $24$ (one map every $2$h for $48$h).

  Section~\ref{ssec:rnn} is dedicated to a quick overview on recurrent neural cells and their convolutional variant.
  In the following sections, we give a general presentation of our prediction schemes.
  % The reader already familiar with RNNs may skip the following section and move to the next one. % ils se sentiront intelligents

  % Convolutional RNN

  \subsection{Convolutional Recurrent neural networks}
  \label{ssec:rnn}
  
  \begin{figure}[!ht]
  \includegraphics[width=\linewidth]{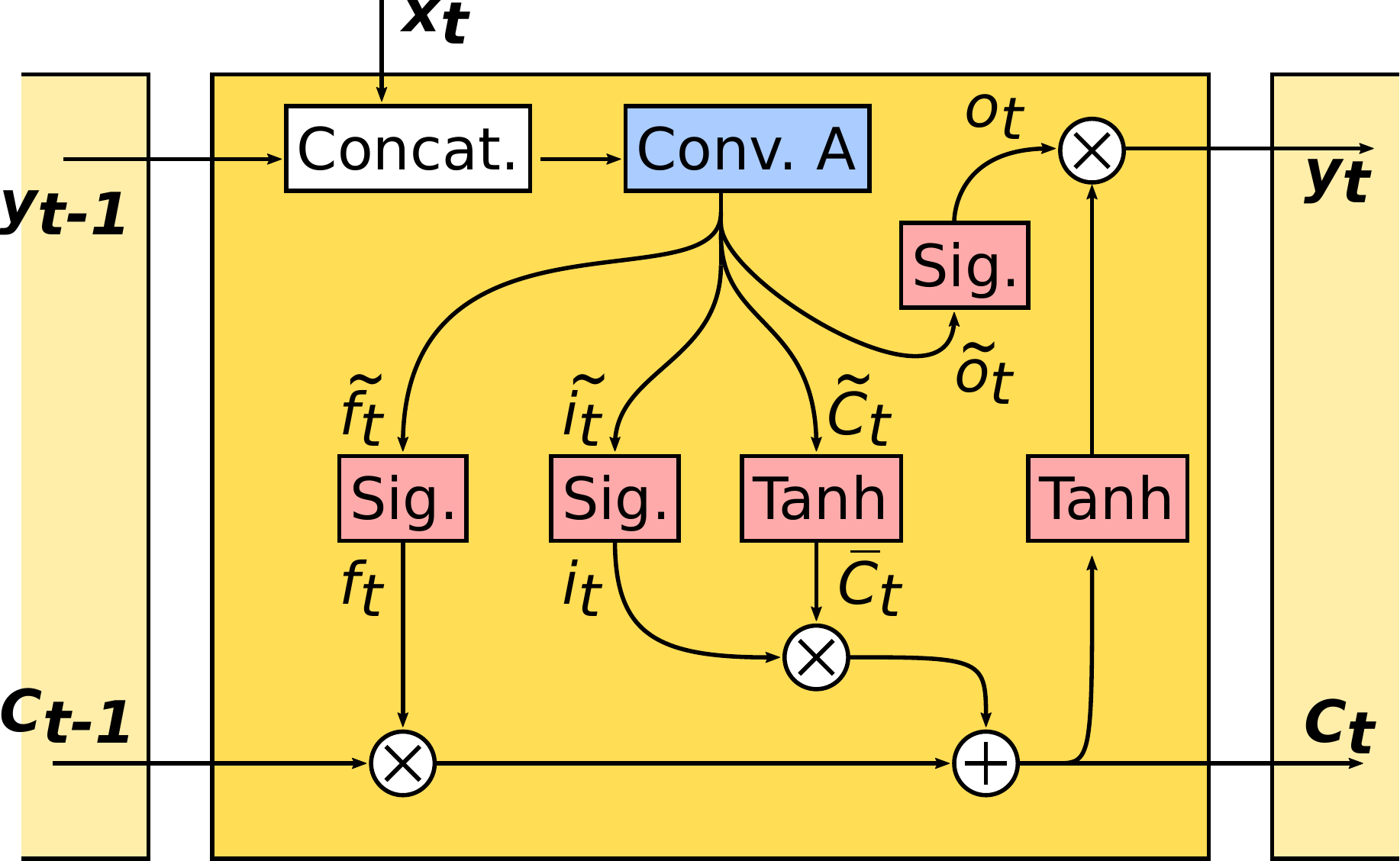}
  
  \textit{Concat. is concatenation, Conv. is convolution layer, Tanh. is the hyperbolic tangent and Sig. is the sigmoid activation.}
  
  \caption{Convolutional LSTM schematic representation.}
  \label{fig:convLSTM}
  \end{figure}
  
  % RNN
  The recurrent neural networks (RNN) \cite{elman1990finding} are known for their ability to exhibit a temporal behavior in the data. 
  In the simplest RNN architecture, the neurons receive simultaneously the input at time $t$ and the previous hidden state at $t-1$. 
  However, these simple RNN have proved to be poorly efficient in memorizing long-term information \cite{LSTM1}.
  Many more efficient architectures exist. 
  The Long Short-Term Memory (LSTM) \cite{LSTM1} involves an internal memory updated at each time step, allowing information to flow further into the network.
  There exist many alternatives to the LSTM that modify the temporal cell architecture, notably the Gated Recurrent Unit (GRU) \cite{gru}.
  
  % CNN
  In this particular paper, we deal with 2D images.
  Images are characterized by strong relations among neighboring pixels. %, e.g. pixels of the same object. % ici l'ionosphere est le seul objet? 
  To deal with the high number of elements and the connectedness of pixels in image processing, neural networks use convolutional layers.
  These layers apply the same learned convolutional kernel (small size) to the pixels and their neighborhood (connectedness).
  This is the key layer of the most popular convolutional neural networks architectures (CNN) such as LeNet \cite{lenet}, VGG \cite{VGG} or ResNet \cite{residual_learning} for image classification, SegNet \cite{badrinarayanan2015segnet} or Unet \cite{unet} for semantic segmentation, or Mask-RCNN \cite{mask_rcnn} for object detection and recognition.
  For a complete review on deep learning and architectures, one could refer to \cite{Goodfellow-et-al-2016}.
  
  Using RNN such as LSTM or other cell architectures would loose the bidimensional information, i.e. the spatial relations between pixels. 
  In order to exploit simultaneously the temporal and spatial information of TEC maps, we use the Convolutional LSTM presented in \cite{xingjian2015convolutional}, also derived in Convolutional GRU in our work. 
  In these units, the usual fully-connected layers are replaced by convolutional layers.
  
  We now detail the convolutional LSTM.
  It has basically the same functioning than the usual LSTM.
  For detailed description of LSTM operation, one can refer to Colah's Blog \footnote{Colah's blog on LSTM: \url{http://colah.github.io/posts/2015-08-Understanding-LSTMs/}}.
  
  Given $x_t$ the input frame at time $t$, $y_{t-1}$ the previous output and $C_{t-1}$ the cell state at $t-1$, the output $y_t$ and next cell state $C_t$ are defined by: 
  
  \begin{eqnarray}
  \left[\tilde{f}_t, \tilde{i}_t, \tilde{C}_t, \tilde{o}_t \right] &=& W_A * \left[x_t, y_{t-1} \right]\\
  \left[f_t, i_t, o_t \right] &=& \left[ \sigma(\tilde{f}_t), \sigma(\tilde{i}_t), \sigma(\tilde{o}_t) \right] \\
  \bar{C}_t  &=& \tanh(\tilde{C}_t) \\
  C_t &=& f_t . C_{t-1} + i_t . \bar{C}_t \\
  y_t &=& o_t . \tanh(C_t)
  \end{eqnarray}
  
  where $[\dots]$ is concatenation, $\sigma$ is the sigmoid function, $W_A * \dots$ denotes the convolution operation of layer $A$ and $.$ is the term-wise multiplication.
  The same graph operation is also presented on Figure~\ref{fig:convLSTM}.

  \subsection{Prediction strategies}

  The challenge of this study is to design a neural network able to handle a specific sequence prediction problem in which both the inputs and targets are a sequence of images (i.e. a sequence of TEC maps).
  The prediction process is achieved by recursively feeding the network with the last prediction and the previous recurrent cell states. 
  Convolutional Neural Networks (CNN) are used to handle the bidimensional structure of the TEC maps.
  Recurrent Neural Networks (RNN) are used to capture the temporal dependencies at different spatial scales.
  The detailed architectures are presented in Section~\ref{sec:networks}.
  Two strategies have been tested: direct prediction and residual prediction.

  \subsubsection{Direct prediction}
  
\begin{figure}[!ht]
  \centering
  \includegraphics[width=0.9\linewidth]{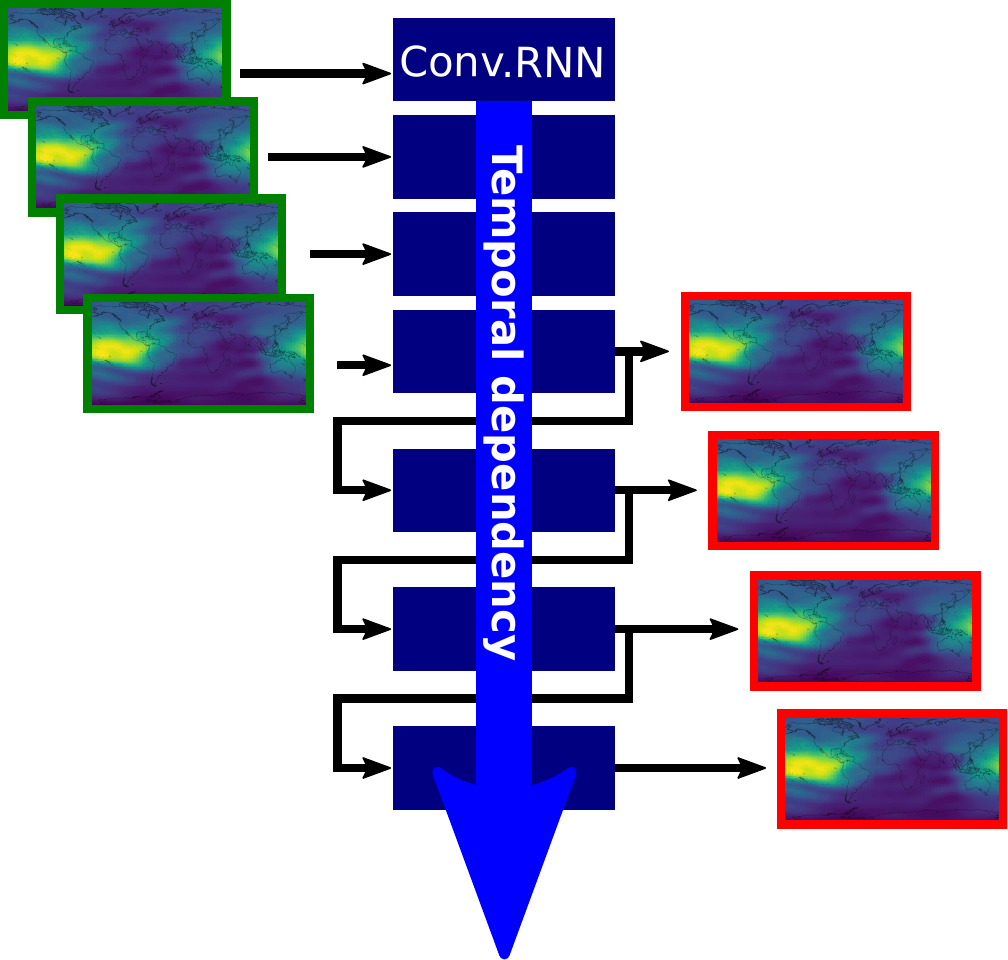}
  \caption{Direct prediction framework.}
  \label{fig:direct}
  \end{figure}

  The first prediction pipeline is a many-to-many prediction framework, represented on Figure~\ref{fig:direct}.
  The input sequence and RNN memory are recursively fed in the network which outputs the next predicted frame.
  The sequence is processed frame by frame in the temporal order by a recurrent convolutional neural network (dark blue box), the temporal information being kept during the iterations of the process (blue arrow).
  When reaching the input frame corresponding to time $t$, the actual prediction begins.
  The prediction for $t+1$ is the network input to predict $t+2$ and so on.
  This architecture can be subject to prediction divergence at first sight.
  By using directly the previous predicted map to forecast the next one, errors in the prediction will somehow cumulate, leading to an error increasing along with time.
  
  \subsubsection{Residual prediction}
  
\begin{figure}[!ht]
  \centering
  \includegraphics[width=0.9\linewidth]{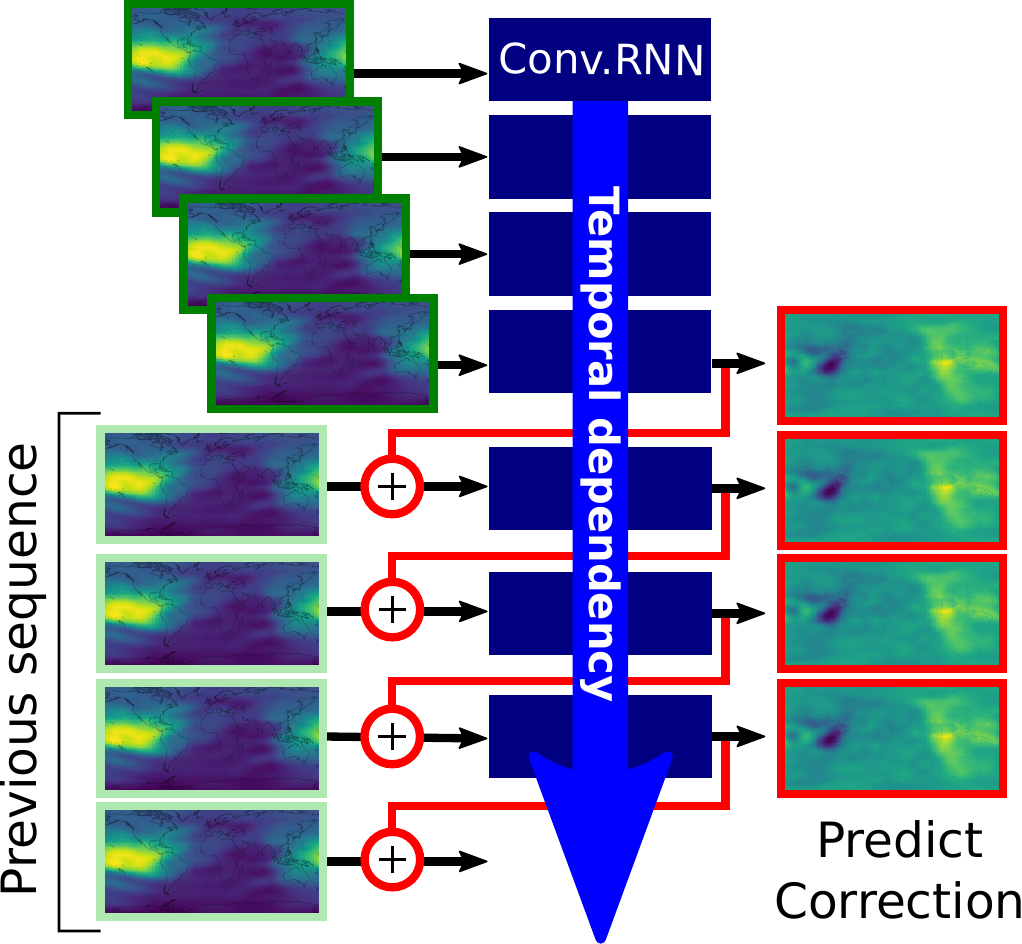}
  \caption{Residual prediction framework.}
  \label{fig:residual}
  \end{figure}

  In order to prevent this possible divergence, we also use a residual prediction, presented in \cite{Cherrier2017}.
  Here the objective is not to predict the next TEC map but a correction, i.e. the difference between the map at $t+1-48h$ and the map at $t+1$.
  The schematic process is shown on Figure~\ref{fig:residual}.
  
  Moreover, to limit the effect of the previous high frequency phenomena in the residual process, we apply a Gaussian blur with $\sigma=3$.
  Not doing so would require the network to memorize the high frequencies at $t-48h$ whereas it is unlikely that they can help in the prediction.

  %%%%%%%%%%%%%%%%%%%%%%%%%%%%%%%%%%%%%%%%%%%%%%%%%%%%%%%%%%%%%%%%%%%%%%%%%%%%%%%%%%%%%%%%%
  %%%%%%%%%%%%%%%%%%%%%%%%%%%%%%%%%%%%%%%%%%%%%%%%%%%%%%%%%%%%%%%%%%%%%%%%%%%%%%%%%%%%%%%%%
  %   _   _ ______ _________          ______  _____  _  __ _____ 
  %  | \ | |  ____|__   __\ \        / / __ \|  __ \| |/ // ____|
  %  |  \| | |__     | |   \ \  /\  / / |  | | |__) | ' /| (___  
  %  | . ` |  __|    | |    \ \/  \/ /| |  | |  _  /|  <  \___ \ 
  %  | |\  | |____   | |     \  /\  / | |__| | | \ \| . \ ____) |
  %  |_| \_|______|  |_|      \/  \/   \____/|_|  \_\_|\_\_____/
  %%%%%%%%%%%%%%%%%%%%%%%%%%%%%%%%%%%%%%%%%%%%%%%%%%%%%%%%%%%%%%%%%%%%%%%%%%%%%%%%%%%%%%%%%
  %%%%%%%%%%%%%%%%%%%%%%%%%%%%%%%%%%%%%%%%%%%%%%%%%%%%%%%%%%%%%%%%%%%%%%%%%%%%%%%%%%%%%%%%%
  \section{Network architectures}
  \label{sec:networks}
  
  %In this section, the challenge is to design a neural network able to handle a specific sequence prediction problem in which both the inputs and targets are a sequence of images (i.e. TEC maps).
  % (Copié collé du début de 3.2)
  
  In this study, we define, use and compare three different network architectures.
  The three of them rely on the convolutional recurrent units presented in the previous section.

  \begin{figure*}[!ht]
  \centering
  \includegraphics[width=\linewidth]{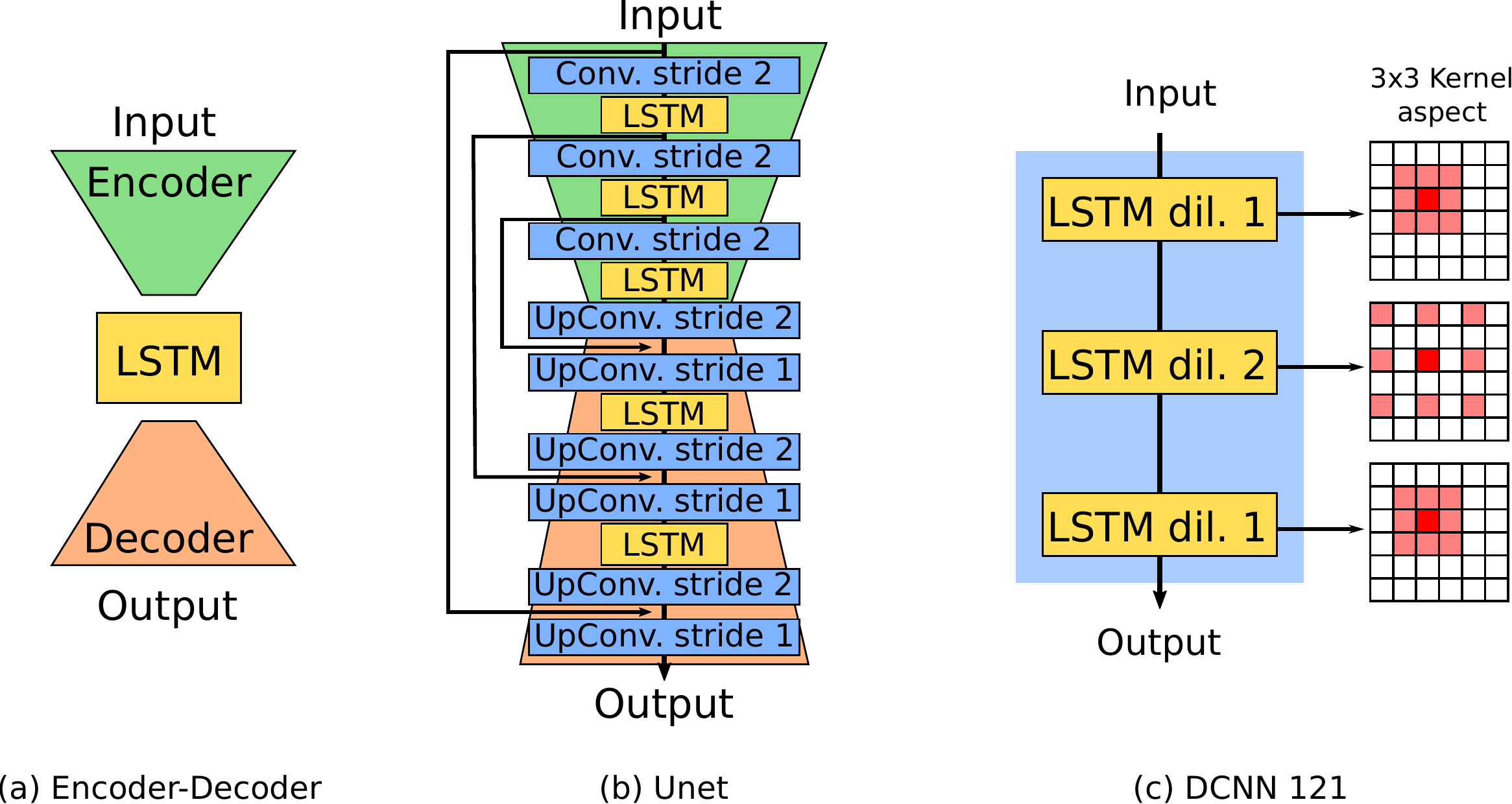}
  \caption{Network used for TEC prediction. From left to right, (a) Network from~\cite{Cherrier2017}, encoder-decoder with a recurrent cell, (b) Recurrent Unet and (c) Dilated Convolution Neural Network $1$-$2$-$1$ inspired from~\cite{Zhang_2017_CVPR}.}
  \label{fig:networks}
  \end{figure*}
  
  \subsection{Encoder - Recurrent unit - Decoder}
  
  The first is the baseline proposed in \cite{Cherrier2017}.
  As presented in Figure~\ref{fig:networks}(a), it has an autoencoder structure \cite{jain2009natural}.
  Convolutional Neural Networks (CNN) are used to handle the spatial structure of the TEC maps.
  A first CNN, the encoder progressively reduces the spatial dimension, creating a coded map, which is the input of the recurrent cell (LSTM).
  Finally, the signal goes through a second CNN, with convolution transpose layers, which increases the signal spatial size back to its original input size.
  
  The encoder has three convolutional layers, eight $3\times3$ kernels each, with stride 2 (for dimension reduction) and a last layer with 4 kernels and stride 1.
  
  % \begin{figure}[!ht]
  % \centering
  % \includegraphics[width=0.35\linewidth]{simple}
  % \caption{Network from~\cite{Cherrier2017}, encoder-decoder with a recurrent cell.}
  % \label{fig:simple}
  % \end{figure}
  
  \subsection{Recurrent Unet}
  
  The next network is an architecture similar to Unet~\cite{unet}.
  Unet was originally designed for semantic segmentation of medical images.
  It also follows an encoder-decoder structure, with progressive spatial dimension reduction before a succession of deconvolutions (decoder).
  
  By using direct convolution transpose layers, the network tries to reconstruct the spatial information it has lost when reducing the spatial size at encoding time.
  It implies that spatial information must be encoded in the signal, often resulting in smooth outputs, details being lost.
  To tackle this problem, Segnet~\cite{badrinarayanan2015segnet} proposes to memorize activations of the max-pooling layers and to reuse them in an unpooling layer, re-injecting the local information at a given scale.
  Unet also retrieves the previous spatial information by concatenating, at a given scale, the feature maps of the encoder and the decoder.
  
  Figure~\ref{fig:networks}(b) shows our recurrent Unet.
  The encoder is a succession of convolutions with stride 2 (reduction of image size by a factor 2) and convolutional LSTM.
  The decoder is an alternation of convolutional blocks (a convolution transpose for dimension augmentation and one for feature mixing with the encoder features) and LSTM layers.
  
  At each layer, as in the previous architecture, the output map number is set to $8$.
  
  % \begin{figure}[!ht]
  % \centering
  % \hspace{-0.6cm}\includegraphics[width=0.5\linewidth]{unet3}
  % \caption{Recurrent Unet.}
  % \label{fig:unet}
  % \end{figure}
  
  \subsection{Recurrent Dilated Network}
  
  Finally, we introduce a third network architecture based on \cite{Zhang_2017_CVPR}.
  For denoising, Zhang et al. \cite{Zhang_2017_CVPR} use a simple, yet efficient network based on 7 convolutions without dimension reduction.
  To increase the receptive field of each pixel, they use a different kernel dilation.
  We show on the right side of Figure~\ref{fig:networks}(c) the aspects of dilated kernels.
  A $3\times 3$ kernel with dilation of 1 is a dense kernel.
  A dilation of 2 corresponds to $5\times 5$ sparse kernel with $0$ every two pixels.
  The resulting dilated convolution neural network (DCNN) has a low memory consumption compared to $5\times 5$ dense kernels.
  
  % \corrA{
  % We derive this network to our purpose.
  % Our architectures is presented on Figure~\ref{fig:networks}(c) and (d).
  % We propose two versions.
  % The first is composed of three recurrent cells with dilation $1$-$2$-$1$.
  % The other is larger with five recurrent cells $1$-$2$-$3$-$2$-$1$ and skip connections similar to Unet.
  % We modified the implementation of the convolutional LSTM cell to accept dilation.
  % For the light version, as with the Unet, the number of output feature maps is set to 8.
  % For the five layer version, it is set to 16.
  % }
  
  \corrA{
  We derive this network to our purpose.
  Our architecture is presented on Figure~\ref{fig:networks}(c).
  It is composed of three recurrent cells with dilation $1$-$2$-$1$.
  We modified the implementation of the convolutional LSTM cell to accept dilation.
  As with the Unet, the number of output feature maps is set to 8.
  }

  \section{Experiments}
  \label{sec:expe}

  \subsection{Note on presented results}
  
  \paragraph*{\textbf{Number of runs}}
  Each training was done 10 times in the same conditions in order to evaluate the performance variation due to training process.
  In the following sections, except when explicitly mentioned, we present the mean performance of the models.
  
  \paragraph*{\textbf{Visualization}}
  For easier understanding of the curves presented in this paper, we apply a temporal filter (mean over a sliding window) of 1 day (12 samples) for the 48h-prediction sequences and 2 days (24 samples) for 2h-prediction.
  By smoothing the curves, we show more clearly the global behavior of each model.
  
  \paragraph*{\textbf{Periodic baseline}}
  Most of the curves presented in this work are relative errors with respect to the periodic baseline.
  It is a simple baseline consisting in using the input sequence as prediction, exploiting the 24-hour periodicity of TEC maps.
  %\comment{Valeur moyenne ?}
  
  \corrA{
  \subsection{Evaluation metric}
  
  In the whole experiment section we use the Root Mean Square (RMS) error to assess the performance of the model.
  The RMS error for a TEC map is given by:
  % \begin{equation}
  % RMS(\mathcal{M}^{t}) =  \frac{1}{|\mathcal{M}^{t}|} \sqrt{ \sum\limits_{i \in \mathcal{M}^{t}} \left(P_{i}^{t}-T_{i}^{t}\right)^2}
  % \label{rmseq}
  % \end{equation}
  \begin{equation}
  RMS(\mathcal{M}^{t}) = \sqrt{ \frac{1}{|\mathcal{M}^{t}|}  \sum\limits_{i \in \mathcal{M}^{t}} \left(P_{i}^{t}-T_{i}^{t}\right)^2}
  \label{rmseq}
  \end{equation}
  where $\mathcal{M}^{t}$ is the set of pixel coordinates corresponding to the TEC map, at $t$, $P$ the predicted map and $T$ the ground-truth map. $i$ indexes the map pixel index and $|\mathcal{M}^{t}|$ is the number of pixels of the map.
  
  We then define the mean RMS for a sequence $S$:
  \begin{equation}
  \overline{RMS}(S) = \frac{1}{|S|}{\sum\limits_{t \in \mathcal{S}} RMS(\mathcal{M}^{t})}
  \end{equation}
  
  and the mean RMS over the test set for frame corresponding to time $t$:
  \begin{equation}
  \overline{RMS}(\mathcal{S}_{test}, t) = \frac{1}{|\mathcal{S}_{test}|}{\sum\limits_{\mathcal{M}^{t} \in \mathcal{S}_{test}} RMS(\mathcal{M}^{t})}
  \end{equation}
  where $\mathcal{S}_{test}$ is the whole test set.
  
  Finally the global mean RMS score is defined by:
  \begin{equation}
  \overline{RMS}(\mathcal{S}_{test}) = \frac{1}{|\mathcal{S}_{test}|}{
  \sum\limits_{S \in \mathcal{S}_{test}} 
  \sum\limits_{t \in S}
  RMS(\mathcal{M}^{t})
  }
  \end{equation}

  }

  \subsection{Training process}
  
  \paragraph*{\textbf{Loss}} For each predicted map, the individual cost function is with respect to the ground truth. 
  \corrA{
  Only the maps in the prediction sequence (images in the future of the element of the input sequence) are considered for penalization.
  }
  Experiments with different loss functions have been conducted in \cite{Cherrier2017}.
  \corrA{
  In table~\ref{tab:loss}, we compare the use of $\ell_2$ and $\ell_1$ losses.
  As in \cite{ma2017sparse} and \cite{Carvalho2018icip}, we get better results with $\ell_1$ loss.
  In the following sections, all results are obtained using $\ell_1$ penalization.
  }
  \paragraph*{\textbf{Optimization}} The network is optimized using a Adam \cite{kingma2014adam} optimizer with learning rate set to $0.001$.
  \corrA{
  We train for 50 epochs.
  For all model, we use a batch size of 16.
  }
  
  \begin{table}
  \caption{ $\ell_2$ vs $\ell_1$, for residual predictions.}
  \label{tab:loss}
  \begin{center}
  \corrA{
  \begin{tabular}{cccc}
  \toprule
  Network & Enc.-Dec. & Unet & DCNN 121 \\
  \midrule
  $\ell_2$& 2.645          & 2.609          &  2.440\\
  $\ell_1$& \textbf{2.626} & \textbf{2.533} &  \textbf{2.423}\\
  \bottomrule
  \end{tabular}
  }
  \end{center}
  \end{table}

  \subsection{Direct vs residual prediction}
  \label{ssec:direct_res}
  
\begin{figure}[!t]
  \centering
  \includegraphics[width=\linewidth]{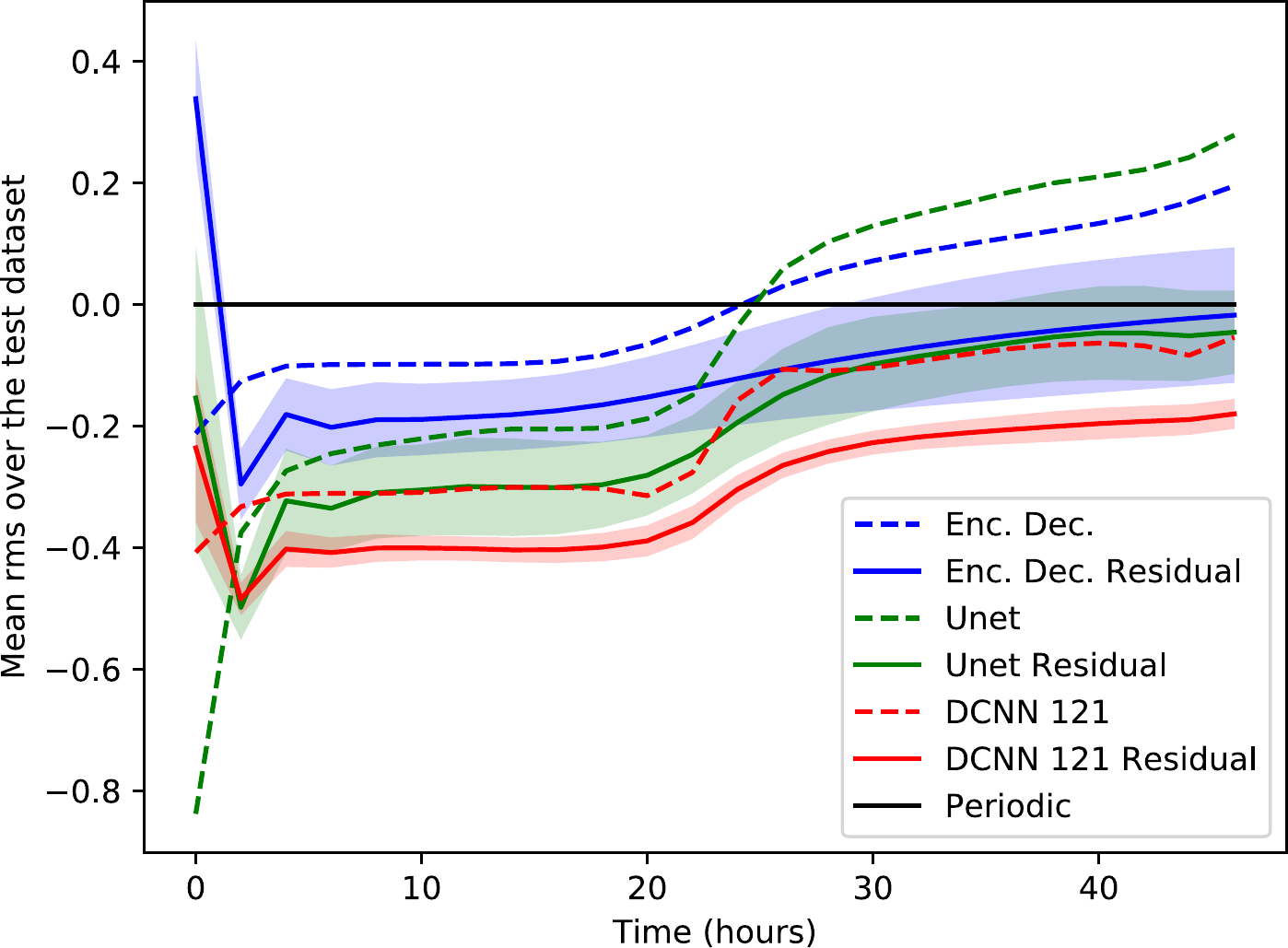}
  \caption{
  \corrA{
  Mean (and standard deviation) error over the dataset for each image of the sequence. Dashed curves are the direct prediction and plain curves are residual.
  For residual, filled area correspond to standard deviation.
  }
  }
  \label{fig:dcnn48hseq}
  \end{figure}

  \corrA{
  We are interested in comparing the two proposed prediction scheme: direct or residual.
  
  \textbf{Global evaluation.~~~}
  Table~\ref{tab:rms} shows the global scores over the test set.
  It presents the RMS along with the corresponding standard deviation and the RMS corresponding to the best run.
  While comparing direct and residual prediction for the each model, we see that both RMS and best runs are performed using the residual scheme.
  Moreover, the standard deviation is always lower in the residual case.
  
  \begin{table}
  \corrA{
  \caption{Mean RMS for the studied networks and baseline, 48h prediction.}
  \label{tab:rms}
  \begin{center}
  \begin{tabular}{llcc}
  \toprule
  Model & Pred. Scheme & Mean $\pm$ std & Best\\
  \midrule
  Periodic & & 2.728 \\
  \midrule
  Enc. Dec. & Direct   & 2.728 $\pm$ 0.088 & 2.611\\
            & Residual & 2.626 $\pm$ 0.077 & 2.539\\
  Unet      & Direct   & 2.668 $\pm$ 0.135 & 2.474\\
            & Residual & 2.533 $\pm$ 0.069 & 2.451\\
  DCNN 121  & Direct   & 2.526 $\pm$ 0.039 & 2.449\\
            & Residual & \textbf{2.423} $\pm$ \textbf{0.024} & \textbf{2.367}\\
  \bottomrule
  \end{tabular}
  \end{center}
  }
  \end{table}
  
  \textbf{Behavior over time though the test set.~~~}
  Figure~\ref{fig:direct_res} presents the prediction curves for the three architectures trained using direct predictions and residual predictions with a 48h forecast horizon.
  The score is the relative $\overline{RMS}(S)$ error compared to the error of the periodic baseline.
  Red plain curve is for direct prediction and blue plain curve is for residual process.
  The scores presented in table~\ref{tab:rms} and the three model observation agree that residual prediction performs best.
  %This is a mean behavior, we can also observe that few sequences are better estimated with direct prediction.
  %This particularly the case when \todo{}

  \textbf{Quality of prediction over the 48h.~~~}
  We can also look at the prediction dimension by computing score each prediction date (between 2h and 48h).
  Figure~\ref{fig:dcnn48hseq} shows the relative error to the periodic baseline with respect to the prediction horizon for models trained on 48h prediction sequences (we discuss in Section~\ref{ssec:horizon} the effect of training for different prediction horizons).
  We use the $\overline{RMS}(\mathcal{S}_{test}, t)$ for each $t$-map of the predicted sequence.
  We first observe that all curves are increasing, showing that prediction becomes more uncertain with time.
  The dashed curves (direct scheme) are almost always above the plain one (residual) except for the first prediction.
  
  \textbf{Interpretation.~~~}
  The direct prediction scheme imposes that the temporal dependencies are only supported by the recurrent cells which is clearly subject to error accumulation along with time.
  On the contrary, for the residual process, injecting the previous TEC maps in the network is safeguard in order to maintain a good prediction.
  It prevents the network from learning to transmit obvious 24h periodic phenomena that are already brought by the blurred sequence.
  The network needs only to learn a correction to a relatively good prediction map while in the direct prediction process, the network learns both correction and 24h periodicities.
  
  However the choice of using residual correction is not always the best one.
  If the prediction horizon is small (e.g. one image corresponding to a 2h forecast ahead of present time), a direct prediction performs better.
  To our understanding, the direct prediction imposes a continuous prediction from one frame to another, which leads to a good first prediction (a small derivation of the last sequence frame).
  On the contrary, the residual forecasting, while constraining the network and being very useful for further prediction, introduces a discontinuity between the last input frame and the first prediction, which is the sum of the predicted correction and the blurred 24h-earlier frame.

  \begin{figure}[!ht]
  \centering
  \includegraphics[width=\linewidth]{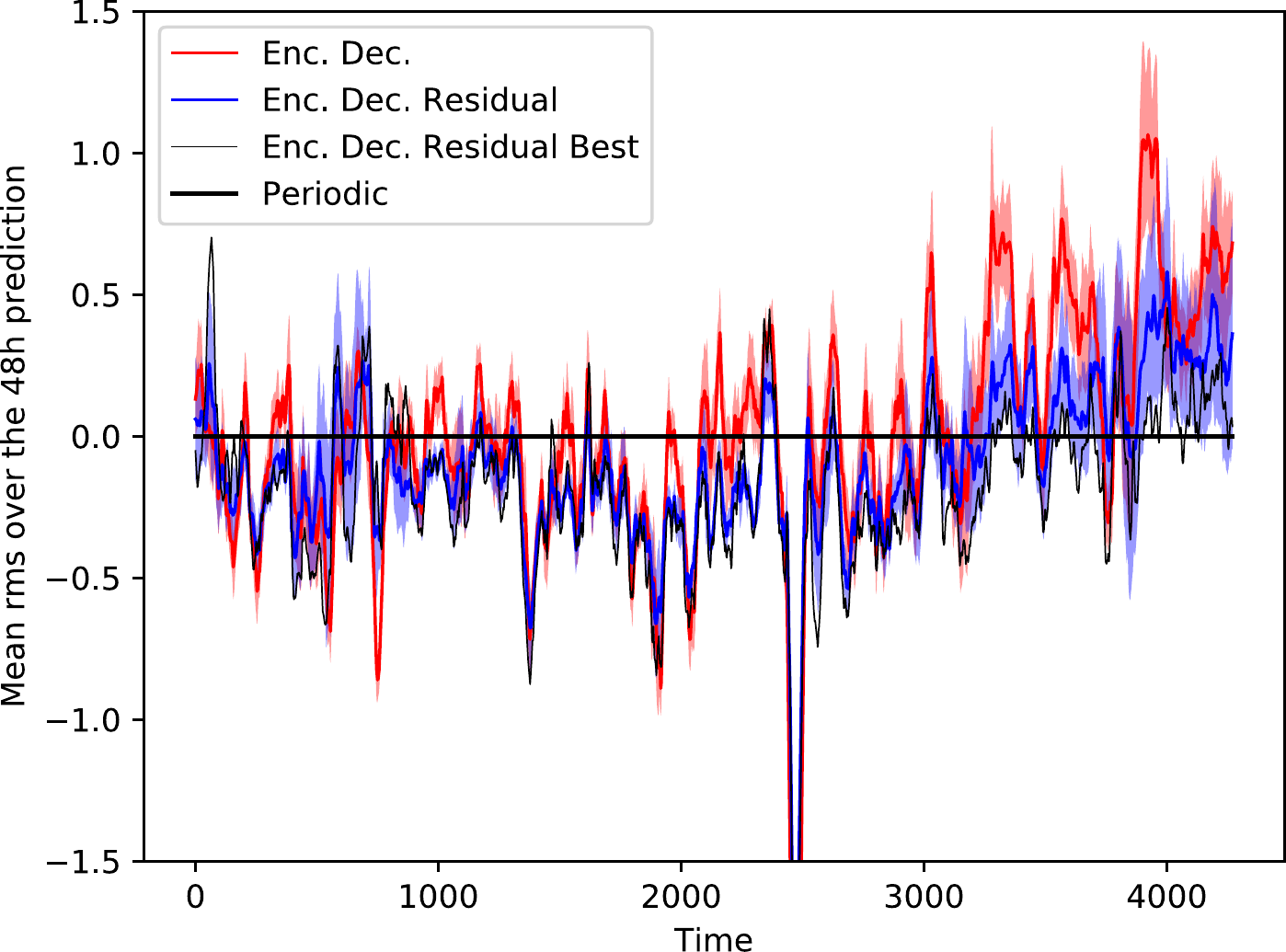}\\
  (a) Simple Encoder-Decoder\\
  \vspace{0.2cm}
  \includegraphics[width=\linewidth]{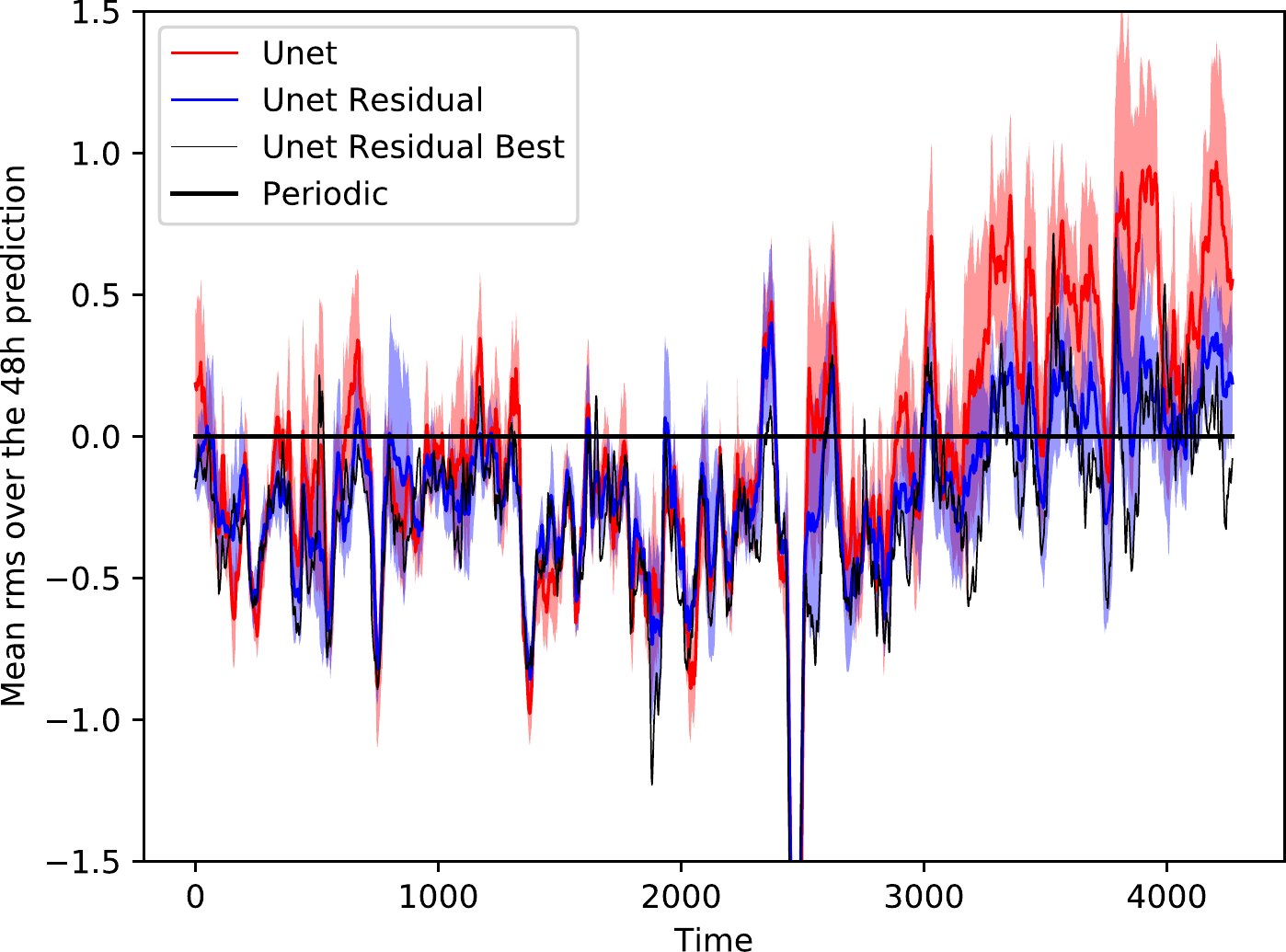}\\
  (b) Unet\\
  \vspace{0.2cm}
  \includegraphics[width=\linewidth]{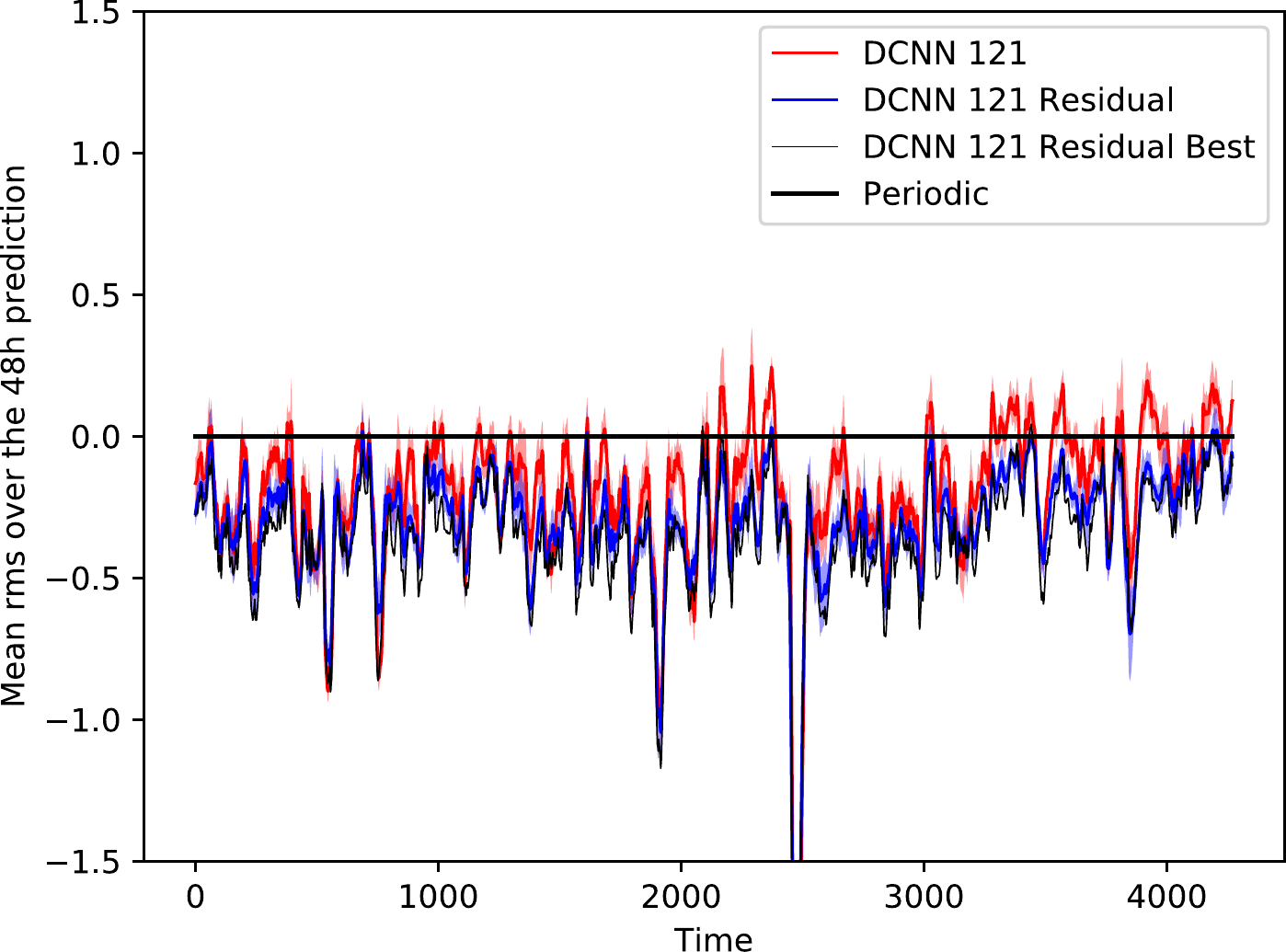}\\
  (c) DCNN
  \caption{48h mean prediction (and standard deviation) error relative to periodic baseline.}
  \label{fig:direct_res}
  \end{figure}

  \corrA{
  Moreover, the step increase of the RMS on figure~\ref{fig:dcnn48hseq} shows that the networks in both direct and residual modes fail to deal correctly with the 24h periodicity.
  In direct mode, the cyclic behavior needs to be well propagated and memorized through the network.
  In residual mode, the blurred sequence comes from the input sequence, i.e. between 24h and 48h this sequence is 
  %more outdated
  less valid 
  than between 2h and 24h.
  }
  
  We can also compare the difference of the two optimization processes with respect to the network architecture.
  On the one hand, the encoder-decoder from~\cite{Cherrier2017} performs poorly with direct prediction.
  The reason is that it contains only one recurrent module, after the encoder: first, the information compression induced by the encoder may lead to information loss and then the only LSTM cell is not sufficient to transmit the complete information needed for a good prediction.
  On the other hand, the Unet and the DCNN proposed in this paper have more similar performances with both forecasting schemes.
  They both contain more recurrent units than in~\cite{Cherrier2017}.
  It allows to more temporal information to flow from frame to frame.
  However the difference between processes still exists and it shows that even these architectures do not succeed in synthesizing good 48h sequences without injecting TEC image priors.
  }

  \subsection{Model comparison}
  
  \corrA{
  In this section, we compare the three presented architectures.
  
  \textbf{Performances.~~~}
  First global RMS scores presented in table~\ref{tab:rms} favors the DCNN architectures both in term of RMS, standard deviation and best run.
  
  The visible interpretation of the plotted curves is congruent with these observations.
  According to Figure~\ref{fig:direct_res} (blue curves), the simple encoder-decoder from \cite{Cherrier2017} is less accurate than Unet and DCNN.
  However, Unet prediction is very oscillating, producing an alternation of good and bad predictions, often even worse than the periodic baseline contrarily to DCNN.
  
  }
  %In this section, we are interested in comparing the networks.
  % According to Figure~\ref{fig:direct_res} (blue curves), the simple encoder-decoder from \cite{Cherrier2017} is less accurate than Unet and DCNN.
  % However, Unet prediction is very oscillating, producing an alternation of good and bad predictions, often even worse than the periodic baseline.
  % On the contrary DCNN and \cite{Cherrier2017} are much more stable (lower standard deviation).
  
  % In Table~\ref{tab:rms}, we aggregate the predictions errors to a single global score $RMS_{global}$ defined by:
  % \begin{equation}
  % RMS_{global} = \frac{1}{|\mathcal{S}_{test}| \; |S|}{\sum\limits_{S \in \mathcal{S}_{test}} \sum\limits_{\mathcal{M}^{t} \in S} RMS(\mathcal{M}^{t})}
  % \end{equation}
  
  %The table presents the mean score of each model with direct prediction, residual prediction and also the RMS score of the best model obtained in training.
  %The indicator for mean results are congruent with the comments on the curves.
  %With a performance enhanced by approximately $5\%$, residual prediction performs significantly better than direct prediction.
  
  The networks introduced in this paper (Unet and DCNN) outperform the network from~\cite{Cherrier2017}.
  To our interpretation, this improvement comes from the higher interdependence between recurrent maps.
  Our networks include several recurrent units against one in \cite{Cherrier2017}.
  The temporal behavior is captured at different spatial scales. 
  Particularly, details do not suffer from the high compression rate operated by the encoder in \cite{Cherrier2017}.
  We also note that not only DCNN better generalizes to test data that is temporally distant from the train set, but it also reaches a better overall score than Unet.
  
  %If we now consider the best trained network, Unet places first, closely followed by DCNN.
  %However, Figure~\ref{fig:direct_res} (black dashed curves) shows that the best run of Unet has the same tendency to diverge with time.
  
  \textbf{Reproducibility and stability over prediction time.~~~}
  \corrA{
  In order to evaluate the variation of performance across several trainings, we plot the standard deviation $\sigma(t)$ around the mean curves of Figure~\ref{fig:direct_res} and Figure~\ref{fig:dcnn48hseq}.
  We observe that the different models do not behave similarly.
  }
  
  % On Figure~\ref{fig:direct_res} and Figure~\ref{fig:dcnn48hseq}, we represent the standard deviation $\sigma(t)$ of the prediction at each time around the mean curve.
  % It expresses the variability of multiple training.
  %On Figures~\ref{fig:direct_res} and \ref{fig:dcnn48hseq}, we observe that the different models do not behave similarly.
  
  The encoder-decoder network has an overall small standard deviation except in the end of the prediction.
  This divergence shows that parts of the trained models fail to generalize to the last quarter of the test dataset.
  Even though the best run (dashed black line) performs fairly well, this result shows that most of time the features learned from the training data are not generic enough.
  When data becomes different from the training set, it fails to generalize.
  
  Unet is the least stable architecture.
  First we observe the same divergence at the end of the testing set and second the standard deviation is very high all along the test set.
  To our opinion, optimizing 5 recurrent layers and multiple conventional convolutions at a time is difficult and might require a larger training set.
  %In most of the runs, the model overfits the training data, failing to generalize on test data.
  \corrA{
  This difference in term of number of parameters is illustrated on table~\ref{tab:nparam}.
  The model from~\cite{Cherrier2017} and the DCNN have almost 4 times less parameters than Unet.
  In most of the runs, the model overfits the training data.
  It is able to predict correctly the beginning of the test set but fails to generalize to the last sequences. This part being more different from the training set due to the large temporal difference.
  }
  
  Finally the dilated convolutional network (DCNN) is the most stable architecture.
  The different runs produce similar performances and more interestingly, the three recurrent units do not diverge with time, i.e. the network learns more generic features.
  On Figure~\ref{fig:dcnn48hseq}, we can also see that the deviation obtained from the residual scheme is spread on the whole sequence.
  On the contrary, with the direct prediction the errors cumulate in time, as underlined before, and the expected performance of training becomes more uncertain.
  
  \begin{table}
  \caption{Number of parameters per architecture.}
  \label{tab:nparam}
  \begin{center}
  \corrA{
  \begin{tabular}{cccc}
  \toprule
  Model                & Enc. Dec & Unet  & DCNN 121 \\
  \midrule
  Number of parameters & 7273     & 28602 & 7592\\
  \bottomrule
  \end{tabular}
  }
  \end{center}
  \end{table}

  \subsection{Training for the prediction horizon of interest}
  \label{ssec:horizon}
  
  \begin{figure}[!ht]
  \centering
   \includegraphics[width=\linewidth]{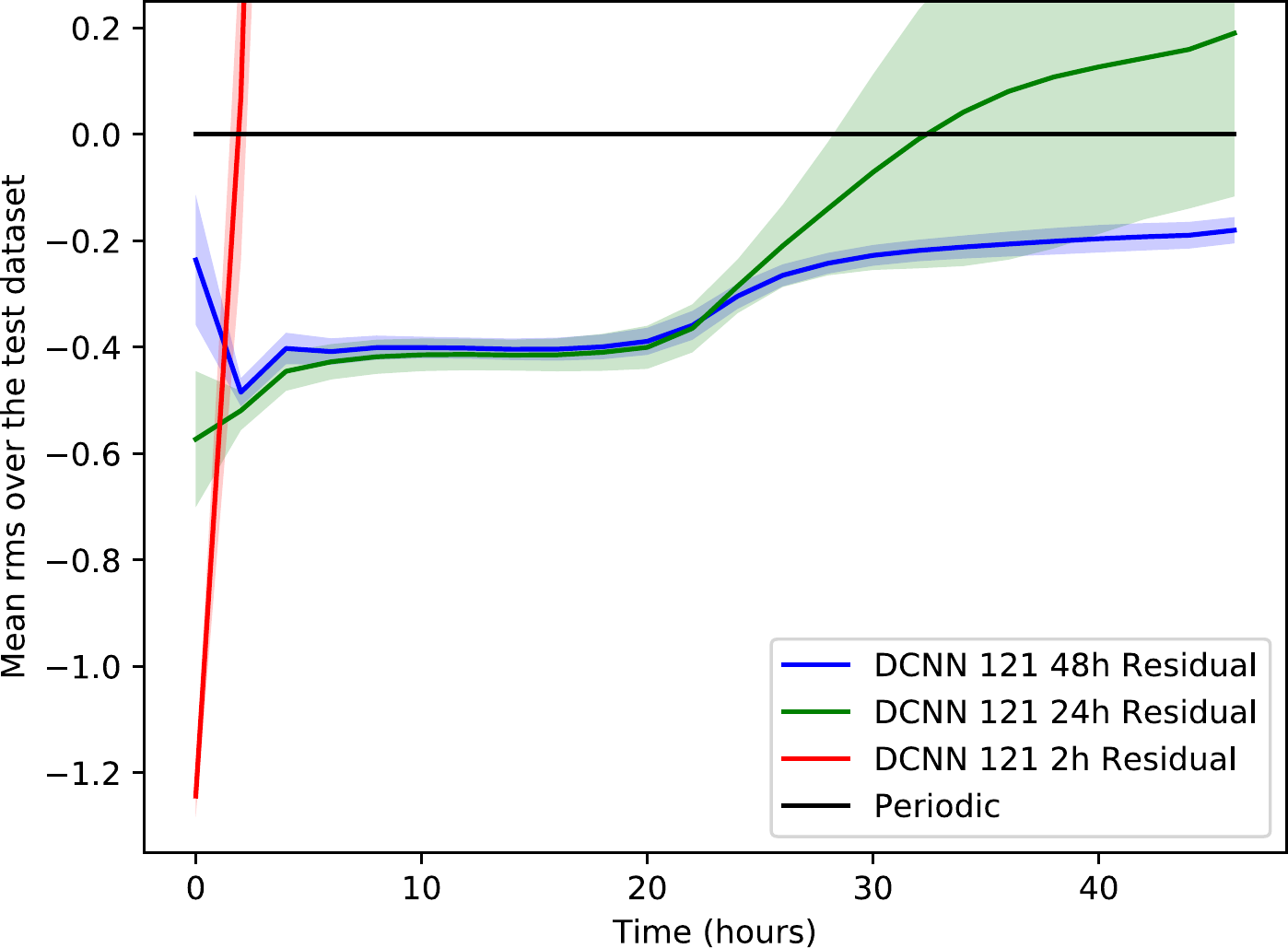}
  \caption{RMS, relative to periodic prediction, over the test set for 2h, 24h and 48h training and prediction horizon.}
  \label{fig:dcnn2hcomp}
  \end{figure}
  
  In the previous sections, we presented networks and optimization for a two days ahead forecast.
  Here we are interested in shorter term prediction, 2h ahead of given data.
  The task is to predict only the TEC map following the last frame of the sequence.
  We compare the effect of training with the same training set but with different objective sequence lengths.
  
  On Figure~\ref{fig:dcnn2hcomp}, we plot the performances of the DCNN 121 network trained to predict sequence of one (2h), twelve (24h) and twenty four (48h) frames.
  \corrA{
  At test time, we let the networks predict 48h and observe the quality of the prediction through the sequence.
  At first glance, we see that the training process is of great influence on the resulting prediction.
  The 2h training is the most efficient for the first frame but fail to generalize to the next frames and diverges quickly.
  Compared to the 48h training, using a 24h horizon improves the prediction on the corresponding time interval but also diverges beyond 12 frames.
  }
  
  % At test time, we evaluate the RMS only on the first frame of the predicted sequence.
  % We used the direct prediction scheme since it performs better for the first frame prediction (Figure~\ref{fig:dcnn48hseq}).
  % The plot confirms the intuitive idea that to predict efficiently 1 frame at 2h, one needs to train the network with that objective.
  
  At training, the network is penalized over the whole sequence and the optimization process spreads the minimization effort on all the frames of the sequence.
  De facto, a network trained to predict only one frame will produce an accurate first prediction, but if we iterate the prediction process, it diverges quickly because it has not learned the temporal dependencies needed to predict more.
  \corrA{
  To a lesser extent, this phenomenon is also at work with 24h compared to 48h.
  }
  
  \subsection{Comparison with literature}

  % \cite{similar6} 2007: very low tec levels this year
  
  % \cite{similar1} 30 minutes
  
  % \cite{similar3} What is the prediction horizon ?

  \begin{table}[htb]\centering
  \caption{Results of previous works.}
  \label{tab:previous_works}
  \begin{tabular}{cc|c|cc|cc}%{@{}lm{3cm}c>{\color{blue}}c@{}} 
  \toprule
  \multicolumn{2}{c|}{Reference} &  RMS (ref) & \multicolumn{2}{c|}{RMS 48h} & \multicolumn{2}{c}{RMS 2h}\\ \midrule 
  \cite{similar6} & %1 station
        $22^\circ$N & 1.45 & \multicolumn{2}{c|}{3.48} & \multicolumn{2}{c}{2.08}\\
  \midrule
  \cite{similar1} & %3 stations 
        $39^\circ$N & \multirow{ 3}{*}{$\leq 2$} &  (1.94) & \multirow{ 3}{*}{2.64} & (1.24) & \multirow{ 3}{*}{1.65}\\ 
  & $30^\circ$N && (2.69) & & (1.71)\\
  & $25^\circ$N && (3.29) & & (2.01)\\
  \midrule
  %\cite{similar3} & \makecell{Global \\ mean TEC} & 3.1 & \multicolumn{2}{c|}{2.46} & \multicolumn{2}{c}{1.53} \\ \bottomrule
  \cite{similar3} & \makecell{Global \\ mean TEC} & 3.1 & \multicolumn{2}{c|}{0.89} & \multicolumn{2}{c}{0.38} \\ \bottomrule
  \end{tabular}
  \end{table}
  
  In Table~\ref{tab:previous_works}, we compare the results for the proposed approach with the results from state-of-the-art models. 
  The presented RMS errors are extracted by selecting the latitude(s) of the station(s) studied in the cited paper.
  \corrA{
  We apply the $\overline{RMS}(\mathcal{S}_{test})$ the TEC maps $\mathcal{M}^{t}$ are reduced to the line of pixels corresponding to the desired latitude. 
  }
  This comparison with previous works on TEC forecasting is only indicative since these works differ by either their prediction horizons or test periods, and since several studies focus on one or a few specific measuring stations instead of producing a worldwide TEC prediction.
  Moreover, \cite{similar6} is a study from 2007, which corresponds to a low activity period of the solar cycle.
  Our train and test sets are drown in a high solar activity period.
  For the second line, \cite{similar1} derive results from 3 ground stations for which we put the individual scores between parenthesis. 
  The score is the mean of the three scores.
  Note that the horizon prediction is $30$ minutes in \cite{similar1} whereas ours is $48$ hours or $2$ hours.
  \corrA{
  We observe that our prediction is better at high latitudes.
  This is simply related to the difficulty of the task depending on the latitude.
  Around the poles, the TEC level is low and does not suffer high variations. On the contrary, around the equator, the TEC values vary a lot both in terms of intensity and shapes, making the estimation task more difficult.
  }
  
  Finally \cite{similar3} predicts the mean TEC level globally.
  For comparison, we apply the following global mean TEC RMS:
  
  \corrA{
  \begin{equation}
  \overline{RMS}_{G}(\mathcal{S}_{test}) = \sqrt{
  \sum\limits_{S \in \mathcal{S}_{test}} 
  \sum\limits_{t \in S}
  \frac{(\overline{P^t} - \overline{T^t})^2}{|\mathcal{S}_{test}|\; |S|}
  }
  \end{equation}
  
  Note that computing directly the mean over the image is not fair due to the distortion of the projection from the globe to the map.
  To compensate this, $\overline{P^t}$ and $\overline{T^t}$ are weighted means where the weight is proportional to the cosine of the pixel latitude.
  
  We observe that even though quality of the prediction depends of the latitude, extrapolating the mean TEC value from the maps gives overall good results.
  The reason is that as we use a regression loss, the model try to reduce the loss for the entire map, which somehow can be interpreted to reducing the global TEC error.
  Moreover compared to~\cite{similar3} we use much more training data as they train their ARMA model on only 180 values.
  
  Nevertheless, we show that the performances of our approach is comparable with state-of-the-art methods, not forgetting that our data-driven approach is global while most of these methods deal with local predictions.
  
  }

  \section{Limitations and future work}
  \label{sec:perspectives}
  
  \corrA{
  A first limitation is the quality of the prediction depends on the forecast horizon the network was trained on.
  Our future work includes creating an architecture able to mix the performance of direct prediction at one frame and the performance of residual prediction for the next frames.
  We will also work on the optimization strategy, optimizing the same network with an increasing prediction length, i.e. as in \cite{flownet2}, starting with relatively easy task and progressively complicating the objective to reach the final prediction horizon.
  
  Second, the continuous data flow is also both an opportunity and a limitation.
  On the one hand, we hope that the increasing amount of data will prevent the bigger networks (i.e. Unet) to overfeat the training data and increase generalization capacity.
  We also consider using larger versions of DCNN, drawing nearer to the original network from~\cite{Zhang_2017_CVPR}.
  On the other hand, the question of iterative learning, i.e. learning from only a subset of the previous state (e.g. the last two years), will arise with growing amount of data.
  Such limitation arise question of the optimization strategy to be adopted in order to update the network weights without downgraded performances.
  Moreover, such strategy, if successful will be an opportunity to deal efficiently with the solar activity cycle, alternating globally low and high TEC levels.
  A model, statically trained on one of these period could lead to poor prediction on the other period.
  
  Next, the main assumption, while being the key to use the proposed approach is that the future states of the ionosphere can be inferred from its previous states.
  We do not take into account the possibility of high-frequency solar perturbations not being captured by the previous TEC maps.
  There are complex dependencies that may not depend on the previous states of the ionosphere.
  As an example solar particles may interfere with the ionosphere \cite{CMEs,solar_activity}.
  The next step will consist in using solar activity parameters as additional inputs to the network.
  Several information sources are to be considered such as multispectral solar images or solar wind parameters \cite{solar_activity}.
  This will lead to rethink the network architecture to include multiple, heterogeneous inputs, with different physical behaviors and dynamics.
  }
  
  % Another limitation is that the quality of the prediction depends on the forecast horizon the network was trained on.
  % It is future work to create an architecture able to mix the performance of direct prediction at one frame and the performance of residual prediction for the next frames.
  % We will also work on the optimization strategy, optimizing the same network with an increasing prediction length, i.e. as in \cite{flownet2}, starting with relatively easy task and progressively complicating the objective to reach the final prediction horizon.
  
  % It is also perspective work to experiment on live learning.
  % The solar activity, and so the TEC measurements, has cycles of several years.
  % Learning from a fixed training set will lead to prediction becoming less accurate with time. 
  % For practical use, a good optimization process would take into account the last prediction sequence with a higher weight and the training set would be updated each time new data is available.

  Finally as an ultimate practical application, a challenge would be to help predicting the positioning correction to apply to GNSS data.
  As an example, native GPS is around 15m and can be improved to 10cm using onerous ground based stations.
  A mid-level correction based on prediction could be of great value.

  %%%%%%%%%%%%%%%%%%%%%%%%%%%%%%%%%%%%%%%%%%%%%%%%%%%%%%%%%%%%%%%%%%%%%%%%%%%%%%%%%%%%%%%%%
  %%%%%%%%%%%%%%%%%%%%%%%%%%%%%%%%%%%%%%%%%%%%%%%%%%%%%%%%%%%%%%%%%%%%%%%%%%%%%%%%%%%%%%%%%
  %    _____ ____  _   _  _____ _     _    _  _____ _____ ____  _   _ 
  %   / ____/ __ \| \ | |/ ____| |   | |  | |/ ____|_   _/ __ \| \ | |
  %  | |   | |  | |  \| | |    | |   | |  | | (___   | || |  | |  \| |
  %  | |   | |  | | . ` | |    | |   | |  | |\___ \  | || |  | | . ` |
  %  | |___| |__| | |\  | |____| |___| |__| |____) |_| || |__| | |\  |
  %   \_____\____/|_| \_|\_____|______\____/|_____/|_____\____/|_| \_|
  %%%%%%%%%%%%%%%%%%%%%%%%%%%%%%%%%%%%%%%%%%%%%%%%%%%%%%%%%%%%%%%%%%%%%%%%%%%%%%%%%%%%%%%%%
  %%%%%%%%%%%%%%%%%%%%%%%%%%%%%%%%%%%%%%%%%%%%%%%%%%%%%%%%%%%%%%%%%%%%%%%%%%%%%%%%%%%%%%%%%
  \section{Conclusion}
  In this paper, we proposed new convolutional neural network architectures for image sequence prediction.
  We applied these networks to global Total Electron Content long range forecasting up to 48h.
  These new networks, based on multiple recurrent neural units, extend the preliminary works of \cite{Cherrier2017}.
  By allowing more temporal information passing, we improved their capacity to predict more complex phenomena.
  %
  %We show that most of the ionosphere behavior can be predicted using its previous states.
  We show that most of the ionosphere behavior can be inferred using its previous states.

  This purely-data-driven approach have been shown to be competitive with state-of-the-art, model-based methods.

  Moreover, the flexibility of neural networks permits to consider future improvements using additional inputs, such as solar imagery.
  
  \section*{Implementation details}
  
  All the presented work was implemented in Python.
  We used Pytorch as deep learning framework.
  The code for dataset generation, training and testing the models is available at 
  \corrA{\url{https://github.com/aboulch/tec_prediction/}}.
  The original convolutional LSTM layer used in our implementation can be found at \url{github.com/rogertrullo/pytorch_convlstm/}.
  \corrA{All experiments were done using a GPU NVIDIA GTX 1070.}
  
  \section*{Acknowledgment}
  
  This work is supported by the DeLTA research programs at ONERA (\url{www.onera.fr/en}).
  DeLTA ( \url{delta-onera.github.io}) is dedicated to experiments on the use of innovative machine learning techniques for aerospace applications from data assimilation to space weather forecasting.
  
  \small
  \bibliographystyle{plain}
  \bibliography{biblio}

  \end{document}